\documentclass[conference]{IEEEtran}
\usepackage{balance}
\IEEEoverridecommandlockouts
\usepackage[utf8]{inputenc}
\usepackage[T1]{fontenc}
\usepackage[english]{babel}
\usepackage{cite}
\usepackage[pdftex]{graphicx}
\usepackage[cmex10]{amsmath}
\usepackage{bbm}
\usepackage{algorithm}
\usepackage{algpseudocode}
\usepackage{array}
\usepackage{booktabs}
\usepackage{multirow}
\usepackage[caption=false,font=normalsize,labelfont=sf,textfont=sf]{subfig}

\begin{document}


\title{Distance Transform Guided Mixup for \\ Alzheimer's Detection}

\author{
  \IEEEauthorblockN{Zobia Batool\IEEEauthorrefmark{1}, Huseyin Ozkan\IEEEauthorrefmark{1}\IEEEauthorrefmark{2}, Erchan Aptoula\IEEEauthorrefmark{1}\IEEEauthorrefmark{2}}
  \IEEEauthorblockA{\IEEEauthorrefmark{1}Faculty of Engineering and Natural Sciences (VPALab), Sabanci University, Istanbul, Türkiye}
  \IEEEauthorblockA{\IEEEauthorrefmark{2}Center of Excellence in Data Analytics (VERIM), Sabanci University, Istanbul, Türkiye\\
    \{zobia.batool, huseyin.ozkan, erchan.aptoula\}@sabanciuniv.edu}
}


%
\maketitle
\IEEEpubid{\makebox[\columnwidth]{\textbf{979-8-3315-6655-5/25/\$31.00 ©2025 IEEE}\hfill}
\hspace{\columnsep}\makebox[\columnwidth]{}}

\begin{abstract}
Alzheimer’s detection efforts aim to develop accurate models for early disease diagnosis. Significant advances have been achieved with convolutional neural networks and vision transformer based approaches. However, medical datasets suffer heavily from class imbalance, variations in imaging protocols, and limited dataset diversity, which hinder model generalization. To overcome these challenges, this study focuses on single-domain generalization by extending the well-known mixup method. The key idea is to compute the distance transform of MRI scans, separate them spatially into multiple layers and then combine layers stemming from distinct samples to produce augmented images. The proposed approach generates diverse data while preserving the brain’s structure. Experimental results show generalization performance improvement across both ADNI and AIBL datasets.
\end{abstract}



%

\IEEEpubidadjcol

\section{Introduction}
Alzheimer's disease (AD) is a progressive syndrome that affects millions of people worldwide. It is caused by complex factors such as aging, genetics, and environment, leading to memory loss and behavioral changes \cite{paper1}. Early and accurate diagnosis of AD is critical for timely intervention and disease management. While MRI scans are widely used in detecting structural brain changes associated with AD, they can significantly vary across datasets due to differences such as scanner hardware and acquisition protocols. Such variations can lead to domain shift, where models trained on one dataset perform poorly on unseen distributions. 

Domain generalization (DG) in Alzheimer’s classification has received limited attention, particularly under the Single DG (SDG) setting, where models are evaluated on unseen target domains with access to a single dataset during training. A disease-driven DG approach \cite{paper2} trained a deep neural network using region-based interpretability, ensuring the model focused on disease-relevant regions via class-wise attention and visual saliency maps. Another study \cite{paper3} proposed ADAPT, a deep-learning model for Alzheimer’s diagnosis that converts 3D brain images into multiple 2D slices and improves model accuracy through a combination of attention, morphology augmentation, and a 2D vision transformer. Alternatively, domain knowledge has been integrated into a ResNet through jointly trained weighted classifiers in \cite{paper4}. Additionally, data augmentation techniques, such as MixUp \cite{paper5} and MixStyle \cite{paper6}, have also been reported for model generalization. MixUp generates augmented training data by linearly interpolating between pairs of images and their labels, while MixStyle perturbs feature statistics to simulate diverse styles. However, these methods may not be optimal for tasks like AD classification, as the random mixing can distort critical disease-specific features. 

To address these challenges, a distance transform guided mixup is proposed, as a structure-aware variant of the mixup technique in order to improve DG in AD detection. It uses a 3D U-Net backbone for brain MRI analysis. Unlike traditional mixup techniques, this approach employs distance transform for region-aware augmentation, enhancing data diversity while preserving structural integrity. This also promotes learning domain-invariant features, essential for generalization across varying imaging conditions. The model is trained on the NACC dataset \cite{paper8} and evaluated on ADNI \cite{paper9} and AIBL \cite{paper10} datasets, where it outperforms baseline models.

\begin{figure*}[t]
    \begin{center}
    \includegraphics[width=1.0\textwidth]{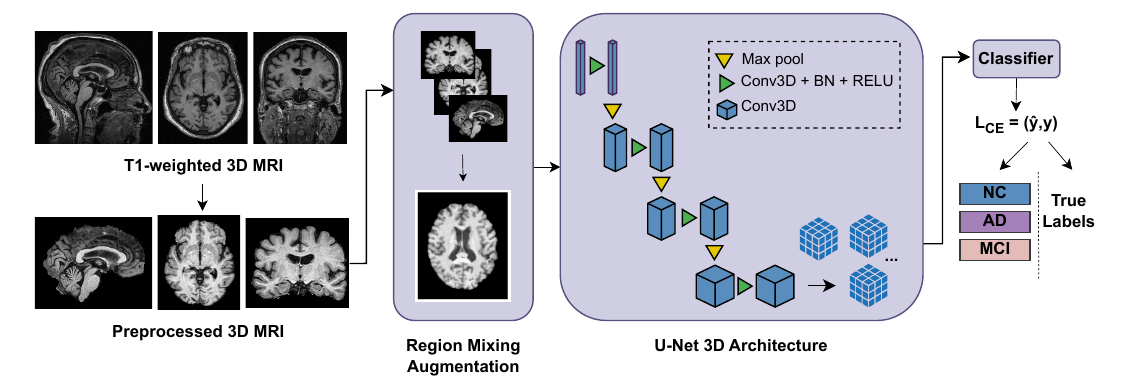}  
    \end{center}
    \caption{Overview of the proposed Alzheimer's disease classification pipeline. 3D MRI scans are preprocessed before applying region mixing augmentation. The augmented images are then processed through a U-Net 3D architecture followed by the classifier.}
    \label{fig:methodology}
\end{figure*}
 
The key contributions of this paper are as follows:
\begin{itemize}
    \item A novel SDG technique is proposed for Alzheimer's classification that enables region-aware mixing without requiring multiple source domains during training.
    \item Structural integrity is maintained in augmented images by leveraging distance transforms and avoiding random blends that could disrupt critical brain regions.
    \item Superior performance is demonstrated on external datasets (ADNI, AIBL) under domain shift conditions, surpassing baseline models.
    \item The proposed augmentation strategy is adaptable for other neuro-degenerative diseases facing similar domain shift challenges.
\end{itemize}

\section{Methodology}
This paper introduces a SDG approach for AD classification, leveraging a 3D U-Net feature extractor with distance transform-based mixup augmentation. The objective is to improve model robustness and generalization while mitigating class imbalance in AD detection from 3D MRI scans.

The training process begins with T1-weighted 3D MRI scans, which are preprocessed for quality enhancement and standardization. The proposed Distance Transform Guided Mixup augmentation technique is then applied, where selected regions from different MRI scans are mixed using distance transform to generate diverse training samples. The augmented images are processed through a U-Net 3D architecture \cite{paper7}, which extracts hierarchical features. These features are fed into a classifier, predicting one of three classes: normal cognition (NC), AD, or mild cognitive impairment (MCI). The model is trained using weighted soft cross-entropy loss to improve classification accuracy and generalization (Fig.~\ref{fig:methodology}).

\subsection{Model Architecture}
The classification framework is built upon a 3D U-Net architecture for feature extraction. The model is initialized with pre-trained weights for chest CT scans\cite{paper7}. To adapt the architecture for the classification task, the decoder is dropped, and the final feature representation is first processed by a global average pooling (GAP) layer followed by two fully connected layers.

\subsection{Distance Transform-Based Mixup Augmentation}
To further improve the model's generalization, for each input MRI scan \( x \), the corresponding distance transform \( D(x) \) is computed offline and stored. Each voxel's value represents its distance to the nearest anatomical boundary. The transform is mathematically defined as:
\begin{equation}
D(p) = \min_{q \in B} \| p - q \|
\end{equation}
where \( D(p) \) is the distance transform value at pixel \( p \), \( B \) is the set of all background pixels and $\| \cdot \| $ is the Euclidean distance. 
\begin{figure}[htbp]
    \begin{center}
    \includegraphics[scale=0.6]{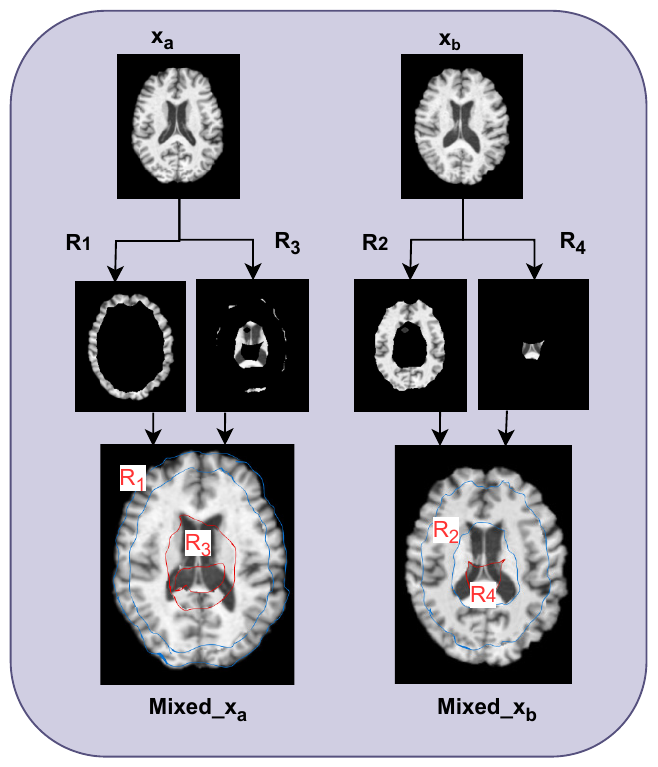}  
    \caption{Overview of the mixing strategy. Given two input MRI scans ($x_a$ and $x_b$), region-wise masks ($R_1, R_2, R_3, R_4$) are extracted to generate mixed samples ($\text{Mixed\_x}_{_a}$ and $\text{Mixed\_x}_{_b}$).}
    \label{fig:transform_methodology}
      \end{center}
\end{figure}
After computing the distance transform, two thresholds \( t_1 \) and \( t_2 \) are set as the minimum and maximum values of the distance transform for each input, ensuring that at least 10\% of the brain structure is preserved in each region. This is done to avoid creating excessively small regions. Given a pair of 3D MRIs \( (x_a, x_b) \), the regions are defined as:
\begin{equation}
    R_1 = \mathbbm{1}(D_a \leq \text{t}_1)
\end{equation}
\begin{equation}
    R_2 = \mathbbm{1}(\text{t}_1 < D_b \leq \text{t}_2) \cdot (1 - R_1)
\end{equation}
\begin{equation}
    R_3 = \mathbbm{1}(\text{t}_1 < D_a \leq \text{t}_2) \cdot (1 - R_1) \cdot (1 - R_2)
\end{equation}
\begin{equation}
    R_4 = \mathbbm{1}(D_b > \text{t}_2) \cdot (1 - R_1) \cdot (1 - R_2) \cdot (1 - R_3)
\end{equation}
where \( \mathbbm{1}(\cdot) \) represents an indicator function that returns 1 if the condition inside is true and 0 otherwise. The variables \( R_1, R_2, R_3, \) and \( R_4 \) are mutually exclusive binary masks, each defining different spatial regions within the image. The terms \( D_a \) and \( D_b \) correspond to the distance transforms of the images \( x_a \) and \( x_b \), respectively. These thresholds divide the MRI into four non-overlapping regions (Fig.~\ref{fig:transform_methodology}). The mixed image is constructed as:
\begin{equation}
    \tilde{x} = (R_1 \cdot x_a) + (R_2 \cdot x_b) + (R_3 \cdot x_a) + (R_4 \cdot x_b)
\end{equation}
where \( R_1, R_2, R_3, R_4 \) represent binary masks corresponding to the different thresholded regions.

\subsection{Label Mixing}
To balance label contributions from the spatially mixed images, the probabilities are calculated based on the pixel count from each region. Given the masks defining the segmented regions, regions \( R_1 \) and \( R_3 \) correspond to the first image \( x_a \) while Regions \( R_2 \) and \( R_4 \) correspond to the second image \( x_b \). The number of pixels assigned from each image is computed as:
\begin{equation}
    P_a = \sum R_1 + \sum R_3
\end{equation}
\begin{equation}
    P_b = \sum R_2 + \sum R_4
\end{equation}
where \( P_a \) and \( P_b \) represent the number of pixels originating from \( x_a \) and \( x_b \), respectively.

The relative contribution of each image to the final mixed sample is determined as:
\begin{equation}
    \alpha_a = \frac{P_a}{P_a + P_b}, \quad \alpha_b = \frac{P_b}{P_a + P_b}
\end{equation}
where \( \alpha_a \) and \( \alpha_b \) represent the proportion of pixels coming from each source image.
Using these proportions, the final mixed label \( \tilde{y} \) is computed as a weighted sum of the labels from both images:
\begin{equation}
    \tilde{y} = \alpha_a \cdot y_a + \alpha_b \cdot y_b
\end{equation}
where $y_a$ and $y_b$ are the original class labels of the input images $x_a$ and $x_b$, respectively. 

A soft cross-entropy loss function is used to handle the soft labels. Additionally, to address the class imbalance in AD classification, class weights \( w_i \) are introduced, ensuring that underrepresented classes contribute more significantly to the loss computation. The weighted soft cross-entropy loss is defined as:
\begin{equation}
    L = - \frac{1}{\sum w_i} \sum_{i=1}^{C} w_i \cdot y_i \cdot \log (\hat{y}_i)
\end{equation}
where \( C \) represents the number of classes,  \( y_i \) represents the true class label, \( \hat{y}_i \) denotes the predicted probability for class \( i \), \( w_i \) is the class weight, computed based on the inverse frequency of each class to mitigate the effects of data imbalance.

\section{Experiments}
To assess the effectiveness of the proposed approach, experiments have been conducted on various baseline methods. The goal is to evaluate model performance in Alzheimer's Disease classification and assess its generalization across different datasets.
\subsection{Datasets}
This study utilizes three publicly available datasets: the National Alzheimer’s Coordinating Center (NACC) \cite{paper8}, the Alzheimer’s Disease Neuroimaging Initiative (ADNI) \cite{paper9}, and the Australian Imaging, Biomarkers, and Lifestyle (AIBL) Study \cite{paper10}. Each dataset includes 3D MRI scans labeled as Normal Control (NC), Mild Cognitive Impairment (MCI), or Alzheimer’s Disease (AD). To ensure consistency, all MRI scans underwent a standardized preprocessing pipeline. This included registration to MNI152 space for anatomical alignment, skull stripping to remove non-brain tissues, and bias field correction to normalize intensity variations. The model was trained and validated on the NACC dataset using an 80/20 split, and its generalization was evaluated on the 80\% of the target datasets, i.e., ADNI and AIBL datasets. Table \ref{tab:demographics} provides demographic details, and participant counts per class for each dataset. 
\begin{table}[H]
    \begin{center}
        \caption{Demographic Characteristics of Participants in NACC, ADNI, and AIBL Datasets.}
        \label{tab:demographics}
        \renewcommand{\arraystretch}{1.2} 
        \setlength{\tabcolsep}{6pt}
        \begin{tabular}{|c|c|c|c|}
            \hline
            \textbf{Dataset} & \textbf{Group (Participants)} & \textbf{Age, years} & \textbf{Gender} \\
            & & mean $\pm$ std & (male count) \\
            \hline
            \multirow{3}{*}{NACC} 
            & NC (n=2524) & 69.8$\pm$ 9.9 & 871 (34.5\%) \\
            & MCI (n=1175) & 74.0 $\pm$ 8.7 & 555 (47.2\%) \\
            & AD (n=948) & 75.0 $\pm$ 9.1 & 431 (45.5\%) \\
            \hline
            \multirow{3}{*}{ADNI} 
            & NC (n=684) & 72.3 $\pm$ 6.9  & 294 (43.0\%) \\
            & MCI (n=572) & 73.8  $\pm$ 7.5 & 337 (58.9\%) \\
            & AD (n=317) & 75.1 $\pm$ 7.7 & 168 (53.0\%) \\
            \hline
            \multirow{3}{*}{AIBL} 
            & NC (n=465) & 72.3  $\pm$ 6.2 & 197 (42.4\%) \\
            & MCI (n=101) & 74.5 $\pm$ 7.2 & 53 (52.5\%) \\
            & AD (n=68) & 73.0 $\pm$ 8.2 & 27  (39.7\%) \\
            \hline
        \end{tabular}
    \end{center}
\end{table}
\subsection{Experimental Settings}
The experiments were conducted on an NVIDIA RTX A6000 GPU (48 GB) using PyTorch 2.3.1. The training ran at 1.76 iterations per second, converging in 9 hours. Due to memory constraints, the initial batch size was 2 but increased to 16 via gradient accumulation. Optimization used SGD with a 0.01 learning rate, 0.9 momentum, and 0.0005 weight decay, with an exponential scheduler reducing the learning rate by 5\% per epoch.

To evaluate the proposed method, experiments were compared against multiple baseline approaches. The MixUp method \cite{paper5} used an interpolation factor ($\alpha$) of 0.3, while RSC \cite{paper11} was configured with a feature dropout rate of 20\% and a background dropout rate of 5\%, with a mixing probability of 0.3. The baseline model \cite{paper7} was a 3D U-Net pre-trained on chest CT scans.

\subsection{Results and Discussion}
This section presents the classification results of the proposed Distance Transform method and baseline approaches on the ADNI and AIBL datasets. Generalization Results for ADNI and AIBL are summarized in Tables \ref{tab:adni_results} and \ref{tab:aibl_results}.

\begin{table}[ht]
\begin{center}
\renewcommand{\arraystretch}{1.2}
\caption{Generalization results on the ADNI dataset.}
\label{tab:adni_results}
\begin{tabular}{|c|cccc|}
\hline
\multirow{2}{*}{Methods} & \multicolumn{4}{c|}{\textbf{ADNI}} \\ \cline{2-5} 
 & ACC(\%) & F1 & SEN & SPE \\ \hline
Baseline \cite{paper7} & 38.04 & 0.359 & 0.359 & 0.679 \\ 
Mixup \cite{paper5} & 48.29 & 0.339 & 0.392 & 0.703 \\ 
RSC \cite{paper11} & 46.14 & 0.407 & 0.410 & 0.713 \\ 
CCSDG \cite{paper12} & 39.55 & 0.396 & 0.419 & 0.700 \\ 
\textbf{Distance Transform (Ours)} & \textbf{48.37} & \textbf{0.460} & \textbf{0.461} & \textbf{0.733} \\ \hline
\end{tabular}
  \end{center}
\end{table}

As shown in Table \ref{tab:adni_results}, on the ADNI dataset, the proposed Distance Transform method outperformed all baselines, achieving the highest accuracy (48.37\%) and F1 score (0.460), along with superior sensitivity (0.461) and, specificity (0.733). This demonstrates its effectiveness in distinguishing between classes, outperforming techniques such as Mixup, RSC, and CCSDG.

\begin{table}[htbp]
\begin{center}
\renewcommand{\arraystretch}{1.2}
\caption{Generalization results on the AIBL dataset}
\label{tab:aibl_results}
\begin{tabular}{|c|cccc|}
\hline
\multirow{2}{*}{Methods} & \multicolumn{4}{c|}{\textbf{AIBL}} \\ \cline{2-5} 
 & ACC(\%) & F1 & SEN & SPE \\ \hline
Baseline \cite{paper7} & 38.50 & 0.338 & 0.392 & 0.699 \\ 
Mixup \cite{paper5} & \textbf{65.42} & 0.382 & 0.382 & 0.721 \\ 
RSC \cite{paper11} & 51.27 & 0.414 & 0.449 & \textbf{0.737} \\ 
CCSDG \cite{paper12} & 40.82 & 0.396 & 0.401 & 0.699 \\ 
\textbf{Distance-Transform (Ours)} & 52.25 & \textbf{0.430} & \textbf{0.454} & 0.726 \\ \hline
\end{tabular}
  \end{center}
\end{table}

As shown in Table \ref{tab:aibl_results}, on the AIBL dataset, the distance transform-based approach yielded promising results, achieving an accuracy of 52.25\%, an F1 score of 0.430, and a sensitivity of 0.454. This demonstrates its robustness in handling domain shifts across datasets. Comparatively, the baseline method achieved only 38.50\% accuracy, while Mixup and RSC provided modest improvements.

The proposed distance transform method showed superior performance across the ADNI and AIBL datasets compared to existing baseline methods. On the ADNI dataset, it outperformed all baselines in terms of accuracy, F1 score, and sensitivity, demonstrating its ability to classify AD, MCI, and NC groups effectively. Similarly, the method achieved competitive results on the AIBL dataset. This reflects the importance of leveraging domain adaptation techniques like Distance Transform to enhance model performance in Alzheimer's Disease classification.

\section{Conclusion}
This paper proposes a distance transform-based method to enhance SDG for Alzheimer’s disease classification. By utilizing distance transforms and region-based mixing strategies, our approach outperformed existing domain generalization techniques, showing the proposed method's robustness and efficiency. Despite these promising results, the computational demands of 3D MRI data, with a batch size of 2, remains a limitation for large-scale applications. Additionally, further exploration is needed to understand the impact of dataset variations (e.g., imaging protocols, scanner differences, demographics) on classification performance. Future work will focus on refining augmentation strategies, improving computational efficiency, and validating our method on additional datasets.
\balance
\section*{Acknowledgment}
This work was supported by The Scientific and Technological Research Council of Turkey (TUBITAK) under Contract 121E452.

\bibliography{references}  

\begin{thebibliography}{10}
\providecommand{\url}[1]{#1}
\csname url@samestyle\endcsname
\providecommand{\newblock}{\relax}
\providecommand{\bibinfo}[2]{#2}
\providecommand{\BIBentrySTDinterwordspacing}{\spaceskip=0pt\relax}
\providecommand{\BIBentryALTinterwordstretchfactor}{4}
\providecommand{\BIBentryALTinterwordspacing}{\spaceskip=\fontdimen2\font plus
\BIBentryALTinterwordstretchfactor\fontdimen3\font minus \fontdimen4\font\relax}
\providecommand{\BIBforeignlanguage}[2]{{%
\expandafter\ifx\csname l@#1\endcsname\relax
\typeout{** WARNING: IEEEtran.bst: No hyphenation pattern has been}%
\typeout{** loaded for the language `#1'. Using the pattern for}%
\typeout{** the default language instead.}%
\else
\language=\csname l@#1\endcsname
\fi
#2}}
\providecommand{\BIBdecl}{\relax}
\BIBdecl

\bibitem{paper1}
J.~Zhang, Y.~Zhang, J.~Wang, J.~Xia, J.~Zhang, and L.~Chen, ``Recent advances in alzheimer’s disease: Mechanisms, clinical trials and new drug development strategies,'' \emph{Signal Transduction and Targeted Therapy}, vol.~9, no.~1, p. 211, Aug. 2024.

\bibitem{paper2}
D.~Lteif, S.~Sreerama, S.~A. Bargal, B.~A. Plummer, R.~Au, and V.~B. Kolachalama, ``Disease-driven domain generalization for neuroimaging-based assessment of alzheimer's disease,'' \emph{Human Brain Mapping}, vol.~45, no.~8, p. e26707, Jun. 2024.

\bibitem{paper3}
Y.~Wang, K.~Chen, Y.~Zhang, and H.~Wang, ``Medtransformer: Accurate ad diagnosis for 3d mri images through 2d vision transformers,'' \emph{arXiv preprint}, Jan. 2024.

\bibitem{paper4}
Y.~Zhou, Y.~Li, F.~Zhou, Y.~Liu, and L.~Tu, ``Learning with domain-knowledge for generalizable prediction of alzheimer’s disease from multi-site structural mri,'' in \emph{Medical Image Computing and Computer Assisted Intervention – MICCAI 2023}, ser. Lecture Notes in Computer Science, H.~G. et~al., Ed.\hskip 1em plus 0.5em minus 0.4em\relax Cham, Switzerland: Springer, 2023, vol. 14224, p. Insert Page Numbers.

\bibitem{paper5}
H.~Zhang, ``Mixup: Beyond empirical risk minimization,'' \emph{arXiv preprint}, 2017.

\bibitem{paper6}
K.~Zhou, Y.~Yang, Y.~Qiao, and T.~Xiang, ``Domain generalization with mixstyle,'' \emph{arXiv preprint}, Apr. 2021.

\bibitem{paper8}
D.~L. Beekly, E.~M. Ramos, G.~van Belle, W.~Deitrich, A.~D. Clark, M.~E. Jacka, and W.~A. Kukull, ``The national alzheimer’s coordinating center (nacc) database: an alzheimer disease database,'' \emph{Alzheimer Disease \& Associated Disorders}, vol.~18, no.~4, pp. 270--277, 2004.

\bibitem{paper9}
R.~C. Petersen, P.~S. Aisen, L.~A. Beckett, M.~C. Donohue, A.~C. Gamst, D.~J. Harvey, C.~R.~J. Jr, W.~J. Jagust, L.~M. Shaw, A.~W. Toga, and J.~Q. Trojanowski, ``Alzheimer’s disease neuroimaging initiative (adni) clinical characterization,'' \emph{Neurology}, vol.~74, no.~3, pp. 201--209, 2010.

\bibitem{paper10}
K.~A. Ellis, A.~I. Bush, D.~Darby, D.~D. Fazio, J.~Foster, P.~Hudson, N.~T. Lautenschlager, N.~Lenzo, R.~N. Martins, R.~Maruff, P.~Masters, and the AIBL Research~Group, ``The australian imaging, biomarkers and lifestyle (aibl) study of aging: methodology and baseline characteristics of 1112 individuals recruited for a longitudinal study of alzheimer’s disease,'' \emph{International Psychogeriatrics}, vol.~21, no.~4, pp. 672--687, 2009.

\bibitem{paper7}
Z.~Zhou, V.~Sodha, M.~M.~R. Siddiquee, R.~Feng, N.~Tajbakhsh, M.~B. Gotway, and J.~Liang, ``Models genesis: Generic autodidactic models for 3d medical image analysis,'' in \emph{Medical Image Computing and Computer-Assisted Intervention – MICCAI 2019}.\hskip 1em plus 0.5em minus 0.4em\relax Cham: Springer International Publishing, 2019, p. 384–393.

\bibitem{paper11}
Z.~Huang, H.~Wang, E.~P. Xing, and D.~Huang, ``Self-challenging improves cross-domain generalization,'' in \emph{Computer Vision–ECCV 2020: 16th European Conference, Glasgow, UK, Aug. 23–28, 2020, Proceedings, Part II 16}.\hskip 1em plus 0.5em minus 0.4em\relax Springer International Publishing, 2020, p. 124–140.

\bibitem{paper12}
S.~Hu, Z.~Liao, and Y.~Xia, ``Devil is in channels: Contrastive single domain generalization for medical image segmentation,'' in \emph{International Conference on Medical Image Computing and Computer-Assisted Intervention}.\hskip 1em plus 0.5em minus 0.4em\relax Springer, 2023, pp. 14--23.

\end{thebibliography}
\end{document}